\documentclass[a4paper,fleqn]{cas-dc}

\usepackage[numbers, sort&compress]{natbib}
\usepackage{float}
\usepackage{microtype}

\def\tsc#1{\csdef{#1}{\textsc{\lowercase{#1}}\xspace}}
\tsc{WGM}
\tsc{QE}
\tsc{EP}
\tsc{PMS}
\tsc{BEC}
\tsc{DE}

\begin{document}
\let\WriteBookmarks\relax
\def\floatpagepagefraction{1}
\def\textpagefraction{.001}

\shorttitle{Hybrid CNN-LSTM network for image-based bean leaf disease classification}

\shortauthors{HJ Rhee and J Akinyemi}

\title [mode = title]{A Resource-Efficient Hybrid CNN-LSTM network for image-based bean leaf disease classification}                      



%
\author[1]{Hye Jin Rhee}[type=editor,
                        auid=000,bioid=1,
                        orcid=0009-0006-7088-0148]

\fnmark[1]



\affiliation[1]{organization={University of York},
    addressline={Heslington}, 
    city={York},
    postcode={YO10 5DD}, 
    country={United Kingdom}}

\author[2]{Joseph Damilola Akinyemi}[style=chinese,
                                    auid=000,bioid=1,
                                    orcid=0000-0003-3121-4231]

\affiliation[2]{organization={University of York},
    addressline={Heslington}, 
    city={York},
    postcode={YO10 5DD}, 
    country={United Kingdom}}

\cormark[1]

\fnmark[2]

\ead{joseph.akinyemi@york.ac.uk}

\cortext[cor1]{Corresponding author}

\fntext[fn1]{Conceptualisation of this study, Methodology, experiments, writing}
\fntext[fn2]{Methodology, experiments, writing and editing}

\nonumnote{In this work, we demonstrate the performance of a lightweight hybrid CNN-LSTM model for bean leaf disease classification from leaf images with impressive results and computational efficiency.
  }

\begin{abstract}
Accurate and resource-efficient automated diagnosis is a cornerstone of modern agricultural expert systems. While Convolutional Neural Networks (CNNs) have established benchmarks in plant pathology, their ability to capture long-range spatial dependencies is often limited by standard pooling layers, and their high memory footprint hinders deployment on portable devices. This paper proposes a lightweight hybrid CNN-LSTM system for bean leaf disease classification. By integrating an LSTM layer to model the spatial-sequential relationships within feature maps, our hybrid architecture achieves a 94.38\% accuracy while maintaining an exceptionally small footprint of 1.86 MB; a 70\% reduction in size compared to traditional CNN-based systems. Furthermore, we provide a systematic evaluation of image augmentation strategies, demonstrating that tailored transformations are superior to generic combinations for maintaining the integrity of diagnostic patterns. Results on the \textit{ibean} dataset confirm that the proposed system achieves state-of-the-art F1 scores of 99.22\% with EfficientNet-B7+LSTM, providing a robust and scalable framework for real-time agricultural decision support in resource-constrained environments. The code and augmented datasets used in this study are publicly available on \href{https://github.com/HJin-R/bean_disease}{this GitHub repo.}
\end{abstract}


\begin{highlights}
\item A hybrid CNN-LSTM architecture that outperforms a baseline CNN on the ibean dataset
\item EfficientNet+LSTM achieves SOTA results with better convergence and generalisation
\item Systematic evaluation of data augmentation techniques for crop disease classification
\item The Hybrid CNN-LSTM model is 70\% smaller than the baseline CNN model
\end{highlights}


\maketitle

\section{Introduction}
Automated plant disease identification is a crucial smart farming system that helps identify real-time crop infections. In common bean (Phaseolus vulgaris L.) plants, prevalent fungal diseases such as bean rust and angular leaf spot compromise the plants' photosynthetic capacity, which can cause structural damage to the plant and extensive loss of yield. For instance, the production loss in Uganda in 2012 caused by bean rust and angular leaf spot was estimated up to 47\% and 55\%, respectively \cite{pamela_severity_2014}. Visual detection remains the essential first step to identifying symptoms, and subsequent microscopic or biochemical component analysis confirms the pathogens. This approach can be expensive and requires processing time \cite{venbrux_current_2023}. Automated disease identification has increasingly considered an attractive means to achieve timely agricultural disease management. \cite{mahlein_plant_2016}

The practical management of common bean crops requires diagnostic tools that are not only accurate but also economically viable for smallholder farmers. While traditional biochemical analysis is a definitive 'gold standard,' the associated costs and logistical delays often render it inaccessible for real-time field management \cite{venbrux_current_2023}. By developing ultra-lightweight architectures that maintain high diagnostic precision while minimising computational overhead, this research provides a potential for Edge-AI deployment. Such systems enable immediate, on-site decision-making, allowing farmers to initiate localised treatments before a minor outbreak scales into the severe yield losses observed in regions like Uganda \cite{pamela_severity_2014}. This shifts the paradigm from reactive agricultural management to a proactive, data-driven approach that is sustainable even in resource-constrained environments.

The development of automated expert systems for agriculture is increasingly essential as global food security faces threats from climate-driven disease outbreaks. Conventional diagnostic processes rely on the availability of human experts, which is often a bottleneck in large-scale farming or remote regions. An effective agricultural expert system must bridge the gap between high-level diagnostic accuracy and operational feasibility on edge devices. This requires a knowledge-based approach to architecture design, where models are not only deep but also resource-efficient. By integrating sequential logic into spatial feature extraction, an expert system can better mimic the human expert's ability to contextualise localised symptoms within the broader structural geometry of a leaf, thereby providing more reliable decision support for crop management.

Artificial intelligence (AI) techniques, such as machine learning (ML) and deep learning (DL), have been instrumental in automated bean leaf disease detection. Traditional ML algorithms such as support vector machines (SVMs) \cite{rahunathan2023} have been used to classify bean leaf disease. DL methods such as CNNs have been even more successful in accurately detecting and classifying diseases \cite{slimani_artificial_2023}. Nonetheless, exploring alternative tools beyond conventional CNNs is in high demand. Leveraging CNN with LSTM has shown encouraging results in various computer vision tasks, such as medical imaging \cite{islam_combined_2020} and human motion recognition \cite{donahue_long-term_2014}. However, this design has not been extensively explored for common bean leaf disease. Furthermore, previous studies that commonly addressed data shortages by implementing data augmentation \cite{onler_feature_2023, elfatimi_beans_2022} have yet to systematically review the impact of various schemes. Notwithstanding, the suitability of a technique is highly subject to various aspects of ML operations.

Our primary experiments involved additional data generation using augmentation techniques and constructing two DL architectures: Bean-CNN (baseline CNN) and Bean-CNN-LSTM (hybrid CNN-LSTM). We trained these models on original and augmented images, and then compared their performance across different augmentation strategies. In additional experiments, we further expanded the data and trained fine-tuned EfficientNet \cite{tan_efficientnet_2019} models using this hybrid design concept. 

The rest of the paper is organised as follows. Section 2 reviews related works. Section 3 describes the methodology employed in our experiments. Sections 4 and 5 present the experimental results and discussion. The final section summarises our findings.

\section{Literature Review}
\subsection{Image-based automated plant disease identification}
With the growing demand for smart farming systems, researchers have proposed techniques for effective image-based plant disease identification. Early research predominantly used traditional ML algorithms combined with feature engineering, as an SVM-based cucumber leaf disease classifier showed the best result with the radial basis function (RBF) \cite{jian_support_2010}. Like many other image classification tasks, adopting CNNs for this problem has significantly improved prediction quality. For example, Lu et al. \cite{lu_identification_2017} showed that their compact CNN model outperformed an SVM-based model in their rice disease classification. Naturally, applying DL methods has become a mainstream approach, as Geetharamani and Pandian \cite{g_identification_2019} proposed a 9-layered CNN trained on the Plant Village dataset \cite{mohanty_using_2016}. More recently, Patil and Manohar proposed an LSTM-CNN tomato leaf disease classifier \cite{patil_plant_2023}. A similar attempt by Devi et al. \cite{devi_plant_2023} showed superior results of their CNN-LSTM plant disease classifier to existing models. \cite{HAQUE2025127743} proposed a Vision transformer (ViT) with triplet multi-head attention to tackle disease detection and achieved a 97.99\% accuracy on plants and apples.

Due to data scarcity, the common bean plant is less studied among agricultural crops \cite{abade_plant_2021}. In most bean disease research, the transfer learning approach using pre-trained models, such as MobileNet \cite{elfatimi_beans_2022} or EfficientNet \cite{singla_deep_2024}, is common. A more recent attempt using YOLO (You only look once) \cite{rodriguez-lira_comparative_2024} was successful in detecting the damage by bean leaf beetles. The customisation of new deep learning architectures is starting to gain momentum, as an article \cite{onler_feature_2023} proposed a combined model using the histogram of oriented gradients (HOGs) and CNN for bean leaf disease identification.

The main drawback of previous attempts is that these complex models have limited usability in real-world situations under hardware constraints \cite{hohman_model_2024}. For real-world deployment, we explored the potential of a hybrid CNN-LSTM as a more efficient deep learning architecture than existing models, and this is rarely attempted for the classification of common bean disease.

\subsection{Image augmentation in plant disease identification}
Data augmentation is commonly applied to image classification problems to improve diversity in training data. The classic model-free approach includes geometric and photometric transformations. Geometric transformations, including rotation, cropping, and flipping, manipulate the geometric primitives of an image. On the other hand, photometric transformations adjust the colour space information, such as RGB (Red-Green-Blue) and HSB (Hue-Saturation-Brightness).
The most popular method in plant disease research is the multi-technique combination due to practicality and general consensus over its effectiveness. This strategy involves applying several transformations to the image data proportionally or randomly. The application of this technique in single-crop \cite{sun_mean-ssd_2021}, and multi-crop classifiers \cite{sym11070939} demonstrated the effectiveness in generating large training datasets. 
Similar to other plant disease studies, this combination technique is a popular approach in bean disease research, typically without reviewing its efficacy. However, various articles \cite{yang_image_2023, taylor2017improvingdeeplearningusing, shorten_khoshgoftaar2019} have suggested that the effectiveness of any augmentation can be highly dependent upon the context, such as the type of task or the model. Therefore, we attempt to review several techniques for our custom model development.

\section{Methodology}
In our initial experiments, we constructed lightweight, custom-designed deep learning architectures built from scratch. For these experiments, we expanded the training samples to three times the original training data by implementing data augmentation techniques. We trained and evaluated our models to contrast their performance to determine which architecture and augmentation techniques were superior to another. 

Our experiments were carried out on a hardware environment with an Intel Xeon CPU running at 2.2GHz and an NVIDIA Tesla 4 GPU. We used OpenCV and Scikit-image packages for augmentation, and TensorFlow for model construction.

\subsection{Data preparation and augmentation for custom lightweight models}
\begin{figure}
    \centering
    \includegraphics[width=0.9\linewidth]{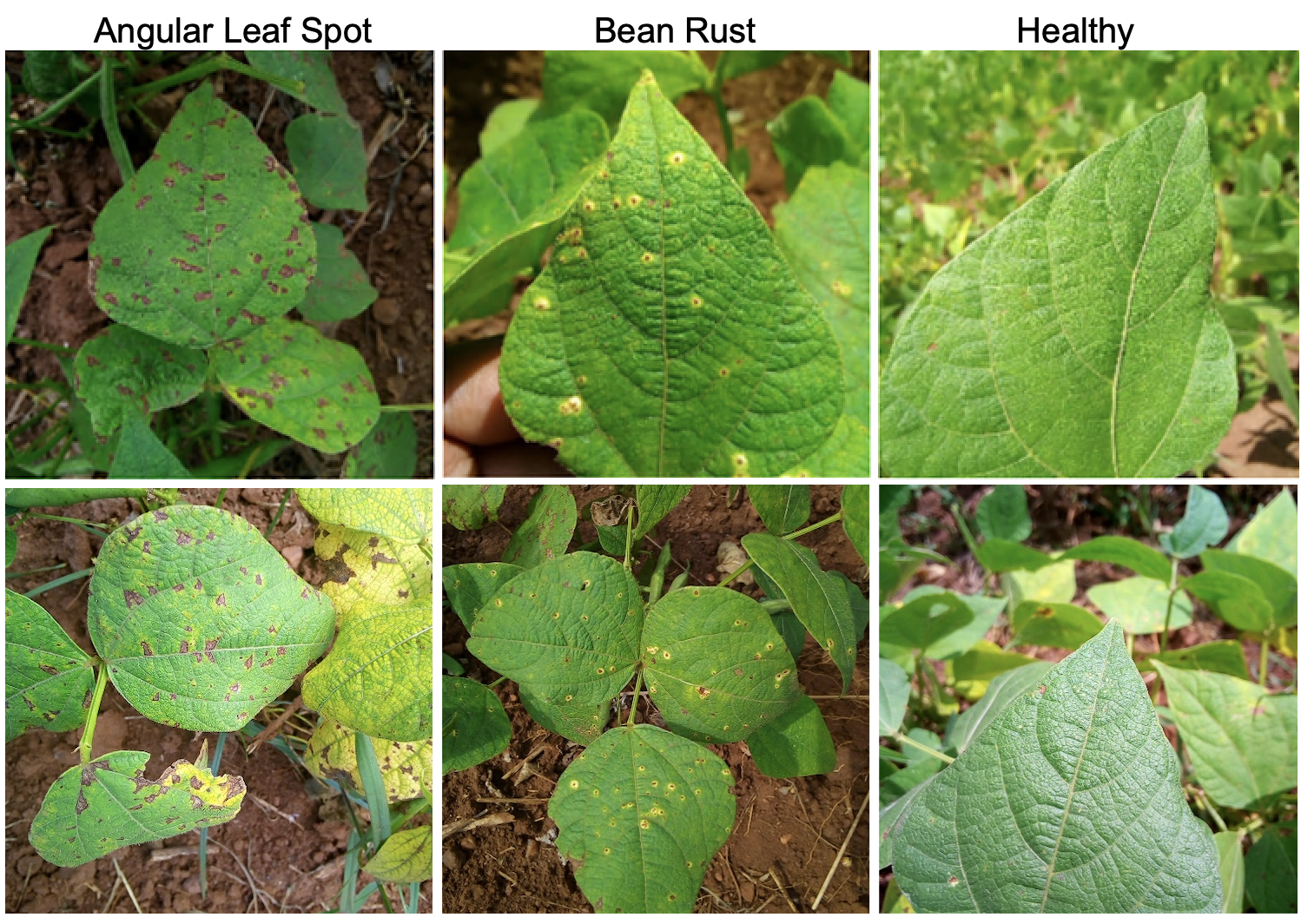}
    \caption{Three classes of bean leaf images in the \textit{ibean} dataset}
    \label{fig:fig1}
\end{figure}

The \textit{ibean} data \cite{noauthor_ai-lab-makerere_2020} is created by Makerere AI Research Lab and contains field-collected bean leaf images showing two disease classes and one healthy class, as shown in Figure \ref{fig:fig1}. Whilst angular leaf spot displays distinctive, angular shaped damage \cite{a_muimba-kankolongo_food_2018}, bean rust exhibits irregular light haloes which may progress to darker blotchy patterns. 

We first combined the original data into one and then split it into 70\% (905 samples) for training, 15\% (195 samples) for validation, and 15\% for testing. In our experiments, the validation/test subsets do not contain augmented data. For augmented training sets (five in total), we applied increased brightness, cropping, flipping, rotation, and random combination. Each augmented training set has 2715 samples with the identical class distribution.

\begin{figure}
    \centering
    \includegraphics[width=0.95\linewidth]{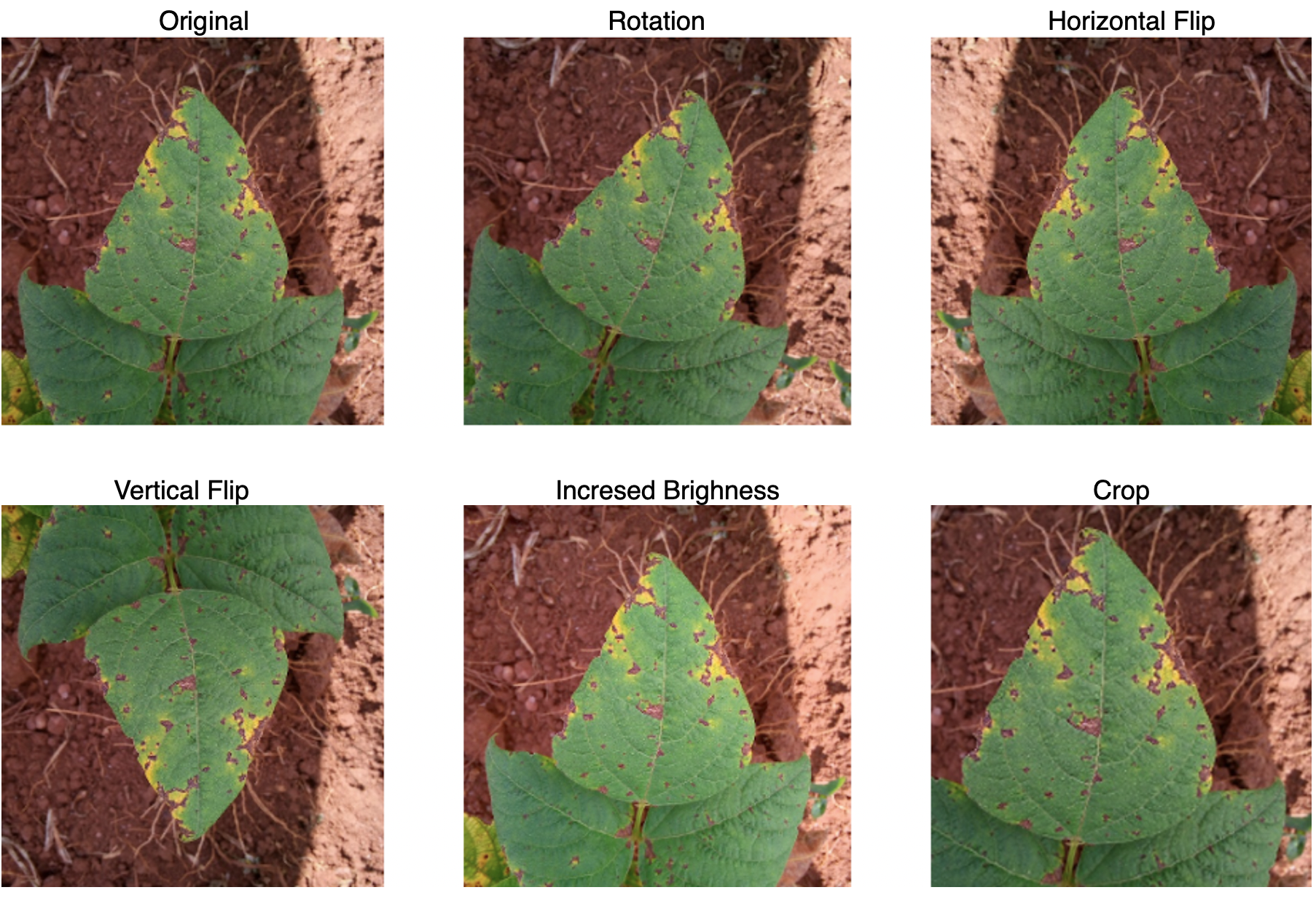}
    \caption{Augmentation effect}
    \label{fig:fig2}
\end{figure}

\subsection{Custom deep learning architecture}
The idea of integrating a CNN and a recurrent neural network (RNN) was first introduced by Deng and Platt \cite{deng_ensemble_2014} for speech recognition. This approach has been considered efficient  \cite{sainath_convolutional_2015} and has been shown to be effective for human-activity recognition\cite{ercolano_combining_nodate}.

The theoretical justification for incorporating an LSTM layer into a computer vision pipeline lies in its ability to model spatial dependencies as a pseudo-temporal sequence. While standard CNNs excel at extracting local hierarchical features through convolutional kernels, they often lack a mechanism to capture long-range relational dependencies across the entire feature map. By treating the flattened feature vectors as a sequence, the LSTM’s gating mechanisms—specifically the input and forget gates—can learn which spatial regions are most indicative of disease clusters versus healthy leaf tissue. In the context of bean leaf pathology, where lesions such as angular leaf spot manifest in sporadic but structurally related patterns, this hybrid approach allows the model to contextualise localised symptoms within the broader leaf geometry, providing a more robust feature representation than isolated fully-connected layers.

\begin{figure}
    \centering
    \includegraphics[width=0.9\linewidth]{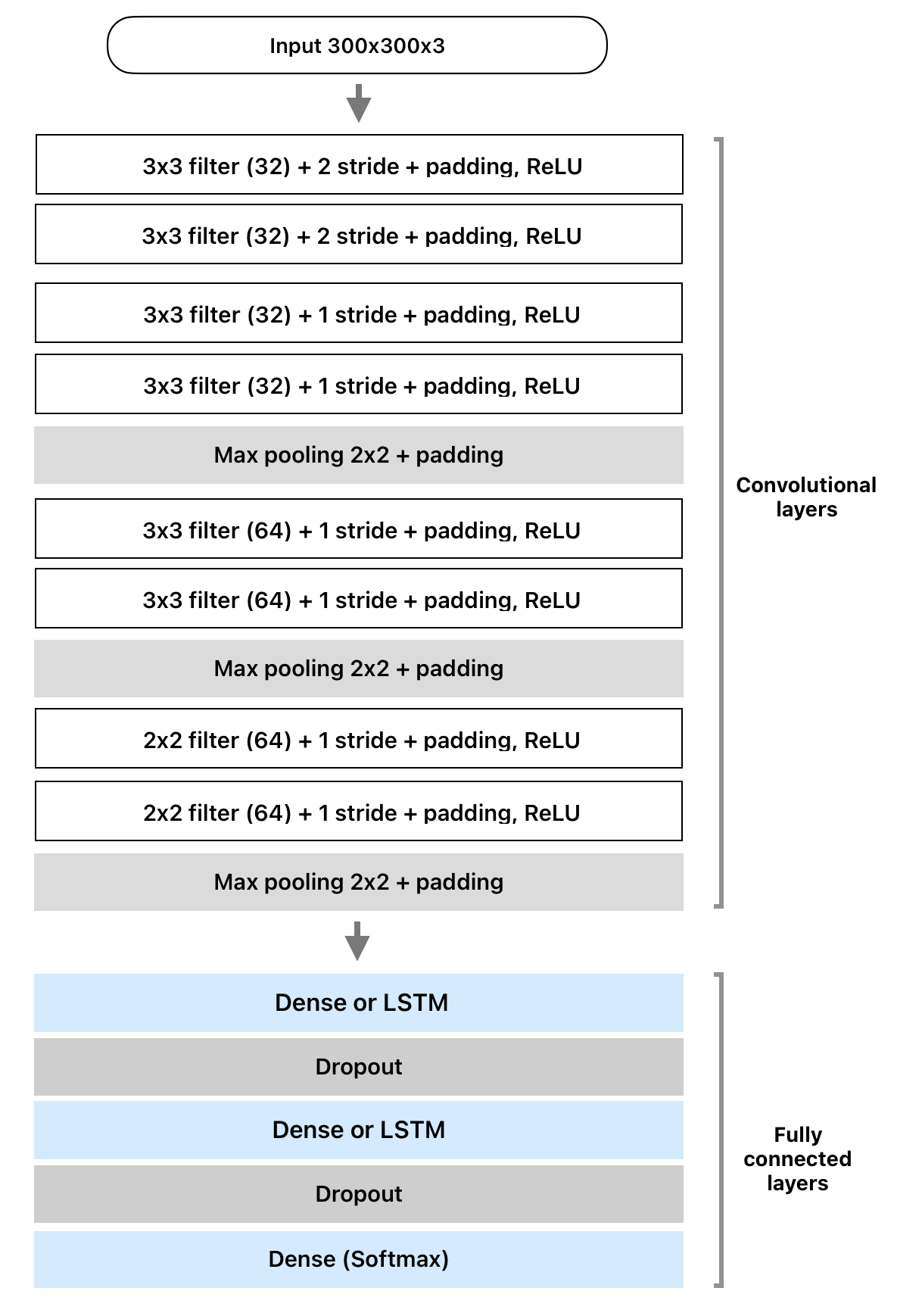}
    \caption{Custom architecture of lightweight models}
    \label{fig:fig3}
\end{figure}

Figure \ref{fig:fig3} presents our proposed model architecture. This is a VGG-inspired \cite{simonyan_very_2015} DL network that offers greater flexibility in modifying the fully connected (FC) layers. We selected this stacked architecture because this has the advantage of capturing intricate patterns in an image over residual networks. \cite{zhang_comparative_2024} Small filters are selected for computational efficiency, and padding parameters are used to minimise the boundary effects. We implemented either the stride parameter or max-pool to reduce the input resolution and improve feature extraction \cite{nixon_feature_2020}. All convolutional layers employ ReLU combined with the Kaiming (He) initialisation to increase non-linearity and ensure more reliable convergence \cite{he_delving_2015}. 

To effectively leverage the LSTM's capability for spatial correlation, we transform the 2D feature maps generated by the final convolutional layer into a structured 1D temporal sequence. Instead of a standard flattening operation, which collapses the spatial hierarchy, the feature maps are reshaped into consecutive time-steps that represent a scan-line or patch-wise traversal of the leaf image. This allows the LSTM's internal gates, specifically the forget ($f_t$) and input ($i_t$) gates, to learn the contextual dependency between adjacent pixel regions. In agriculture, where disease manifestations such as bean rust spread in sporadic and blotchy clusters, the LSTM acts as a secondary filter that identifies the recurring geometric patterns of pathogens across the entire leaf surface, leading to a potentially improved performance gain over baseline CNNs. Hence, whilst the vectors fed into the densely connected neural network (NN) in Bean-CNN were flattened to preserve spatial information, these vectors were sequentially reshaped in time steps for Bean-CNN-LSTM. An LSTM is a modified variant of RNNs \cite{10.1162/neco.1997.9.8.1735} and uses the backpropagation through time (BPTT) technique in its cell-based computations, allowing the gradient-update across multiple time steps. The relevant LSTM's computations are expressed in the following equations. 

\begin{equation}
f_t = \sigma({w_f} \cdot [h_{t-1}, x_t] + b_f)
\end{equation}

\begin{equation}
c'_t = \tanh({w_c} \cdot [h_{t-1}, x_t] + b_c)
\end{equation}

\begin{equation}
i_t = \sigma({w_i} \cdot [h_{t-1}, x_t] + b_i)
\end{equation}

\begin{equation}
c_t = f_t * c_{t_1} + i_t * {c'_t}
\end{equation}

\begin{equation}
o_t = \sigma({w_o} \cdot [h_{t-1}, x_t] + b_o)
\end{equation}

\begin{equation}
h_t = \tanh(c_t) * o_t
\end{equation}

Here, the typical vector equation for the forget gate of an LSTM cell is represented as $f_t$, input gates as $i_t$ and $c'_t$, update gate as $c_t$, output gate as $o_t$, and the hidden state as $h_t$, where $w$ and $b$ denote the weight and bias of the corresponding gate. $x_t$ is the data vector at the current time step $t$ and  $t-1$ indicates the previous time step. $\sigma$ denotes the logistic sigmoid, and  $\tanh$ is the hyperbolic tangent function. 

To examine the quality of the estimated weights, our proposed models use the softmax and sparse categorical cross-entropy loss function. The probability calculated by softmax is integrated with categorical cross-entropy to approximate the prediction error. The following equation describes the computation of this loss function $L$, where $v_k$ is the prediction in $k$ classes for the $j_{th}$ class of $i_{th}$ data point and $y_i$ is the true distribution. 
\begin{equation}
    {L} = -\sum_{i=1}^{k}y_i log(\frac{e^{v_i}}{\sum_{j=1}^{k} e^{v_j}})
\end{equation}

For model evaluation, we collected accuracy, the confusion matrix, the F1 score, and the Matthews correlation coefficient (MCC), for more balanced analysis \cite{tharwat_classification_2021, chicco_advantages_2020}. The confusion matrix represents the instances of the predicted label against the true label. Correctly predicted instances are true positives (TP) and true negatives (TN). In contrast, false positives (FP) and false negatives (FN) represent erroneous predictions. Based on the confusion matrix, F1 and MCC can be calculated as,

\begin{equation}
Precision = \frac{TP}{TP + FP}
\end{equation}

\begin{equation}
Recall = \frac{TP}{TP + FN}
\end{equation}

\begin{equation}
F1 = 2 * \frac{Precision * Recall}{Precision + Recall}
\end{equation}
{\small
\begin{equation}
 MCC = \frac{TP*TN - FP*FN}{\sqrt{(TF+FP)(TP+FN)(TN+FP)(TN+FN)}}
\end{equation}
}

\section{Experiments and Results}
\label{sec:results}

\subsection{Experimental setup for custom lightweight models}
To ensure the best performance and the balance between overfitting and underfitting, we optimised each model's capacity, such as the number of hidden layers and units. Table \ref{tab:tab1} describes the optimal parameter and hyperparameter values that we found during the iterative fine-tuning process for our custom models.

\begin{table}
\caption{\textbf{Configuration of custom lightweight models}}
\label{table}
\setlength{\tabcolsep}{3pt}
\begin{tabular}{|p{110pt}|p{45pt}|p{45pt}|}
\hline
\text{Parameter} &
\textbf{Bean-CNN} &
\textbf{Bean-CNN-LSTM} \\
\hline
No. of hidden units in the $1^{st}$ and $2^{nd}$ layer & 64, 8 & 64, 16 \\
Optimiser & Adam & Adam \\
Learning rate & 1e-3 & 1e-3 \\
Epochs & 40 & 40 \\
Optimiser & Adam & Adam \\
Batch size & 32 & 32 \\
\hline
\end{tabular}
\label{tab:tab1}
\end{table}

\begin{figure}
    \centering
    \includegraphics[width=1\linewidth]{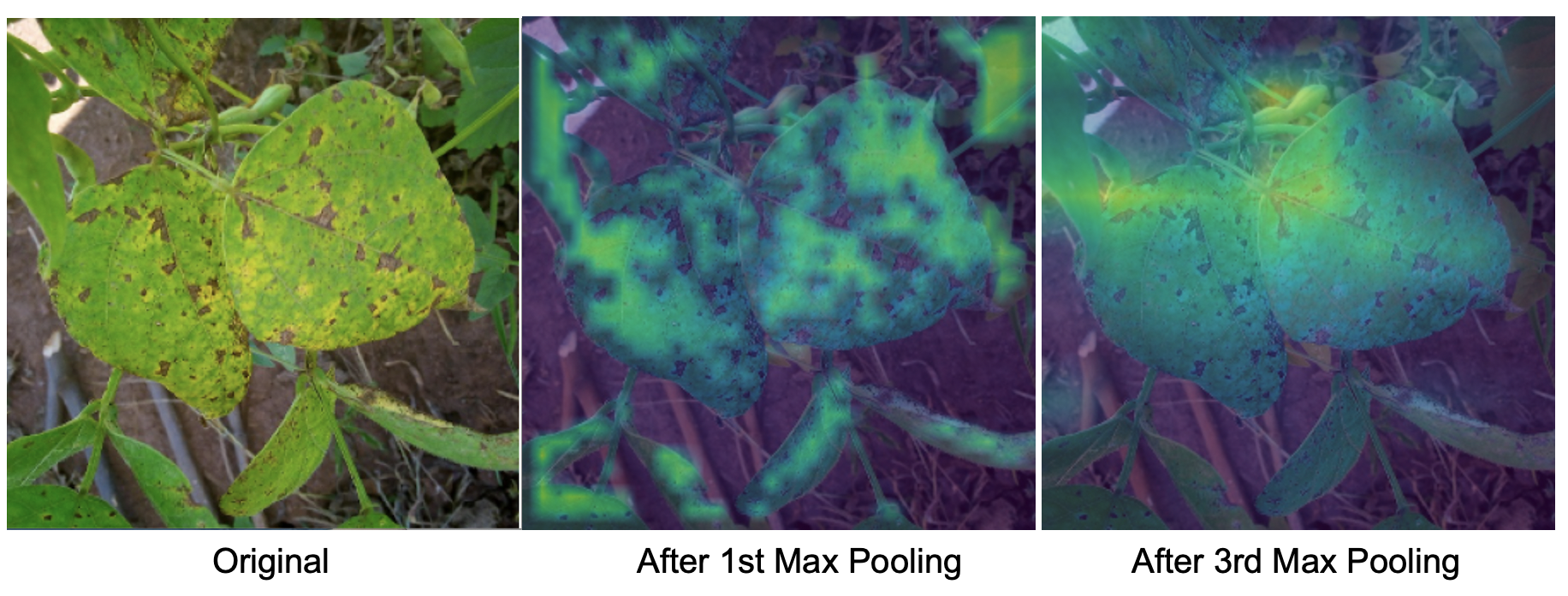}
    \caption{Activation map by custom lightweight model}
    \label{fig:fig4}
\end{figure}

We used Gradient-Weighted Class Activation Mapping (Grad-CAM) \cite{selvaraju_grad-cam_2017} to visualise the features extracted in the convolutional layers. The weight distribution is represented in a heat-map in Figure \ref{fig:fig4} (in the Viridis scale) and shows that the feature extraction by the custom models becomes more specific in deeper layers.

\subsection{Results: Bean-CNN vs. Bean-CNN-LSTM}

\begin{figure}
    \centering
    \includegraphics[width=0.9\linewidth]{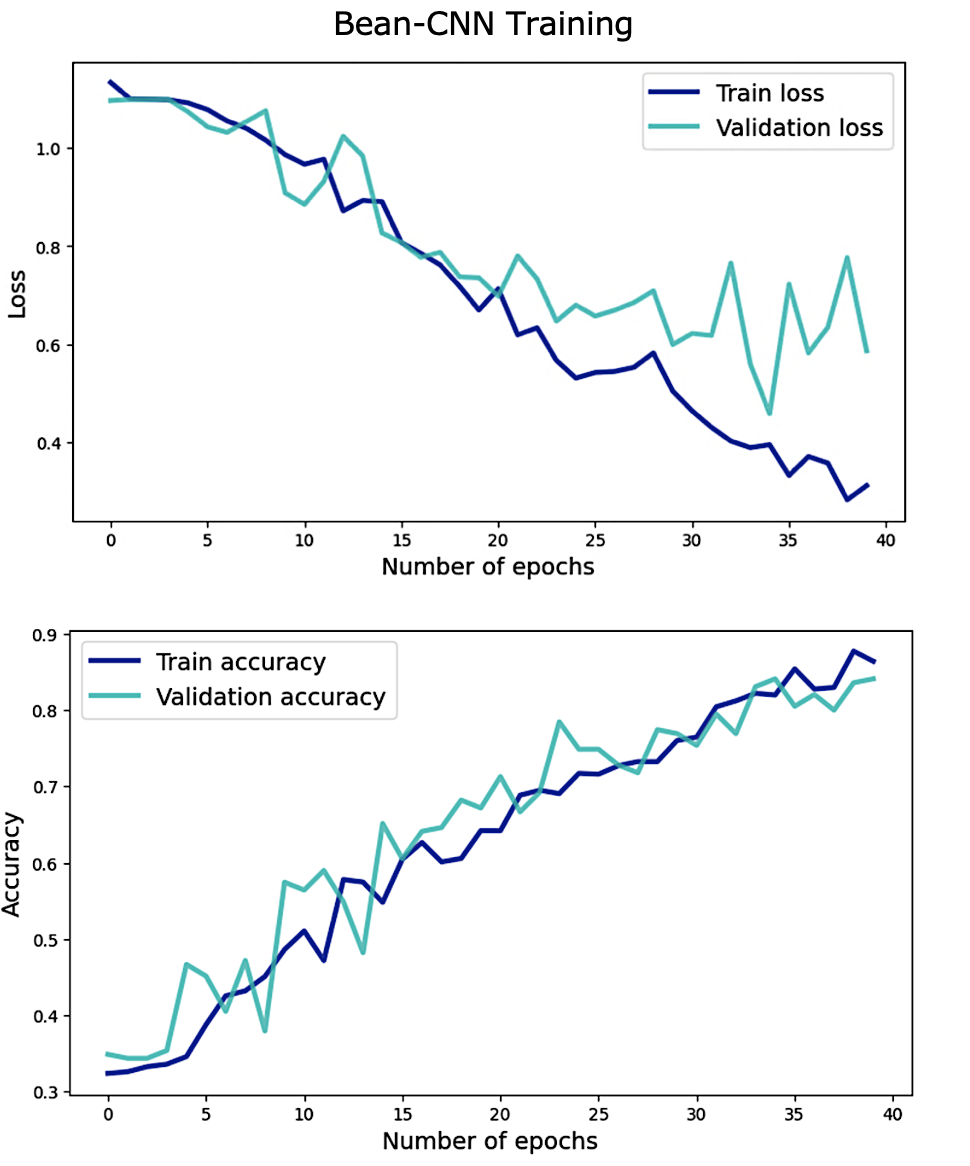}
    \caption{Training Bean-CNN on the original set}
    \label{fig:fig5}
\end{figure}

\begin{figure}
    \centering
    \includegraphics[width=0.9\linewidth]{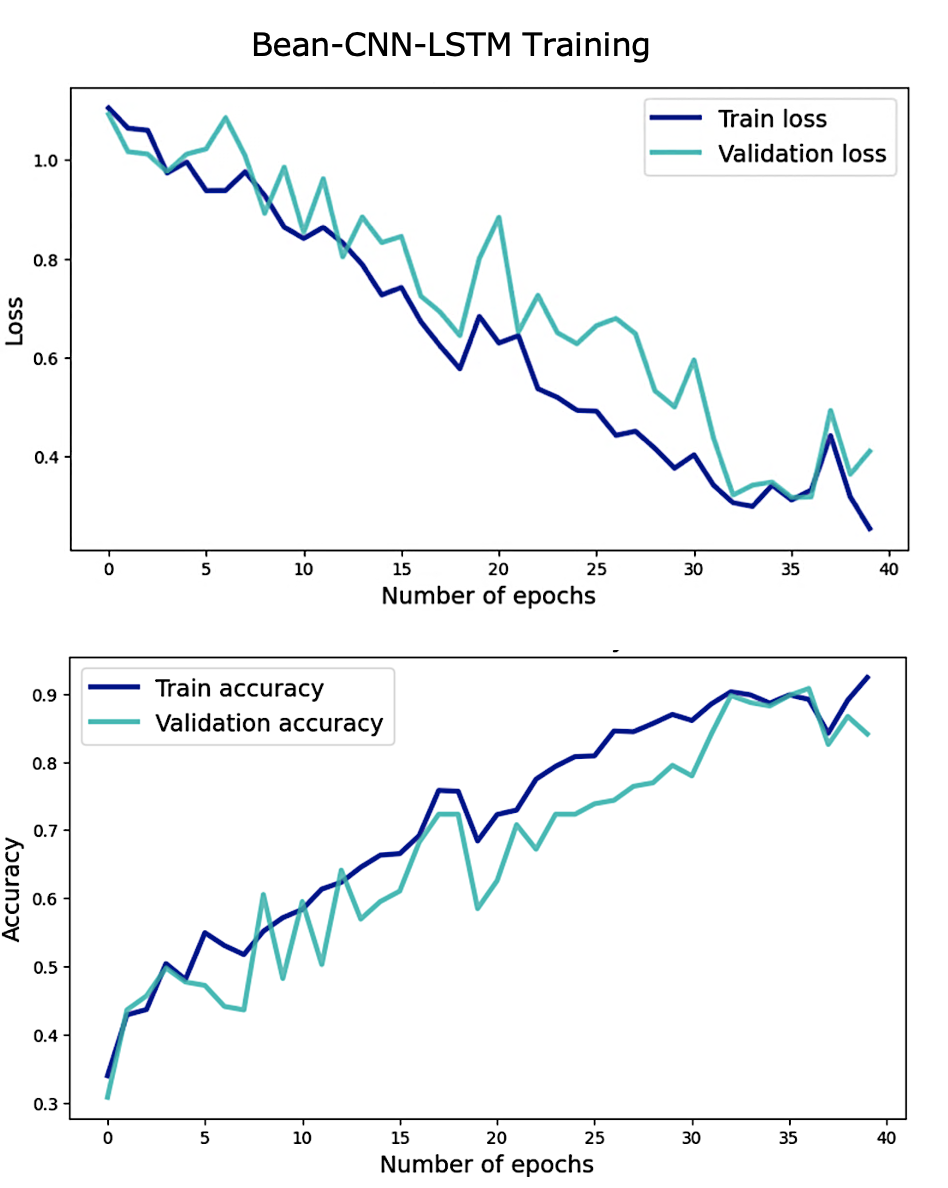}
    \caption{Training Bean-CNN-LSTM on the original set}
    \label{fig:fig6}
\end{figure}

\begin{figure}
    \centering
    \includegraphics[width=0.9\linewidth]{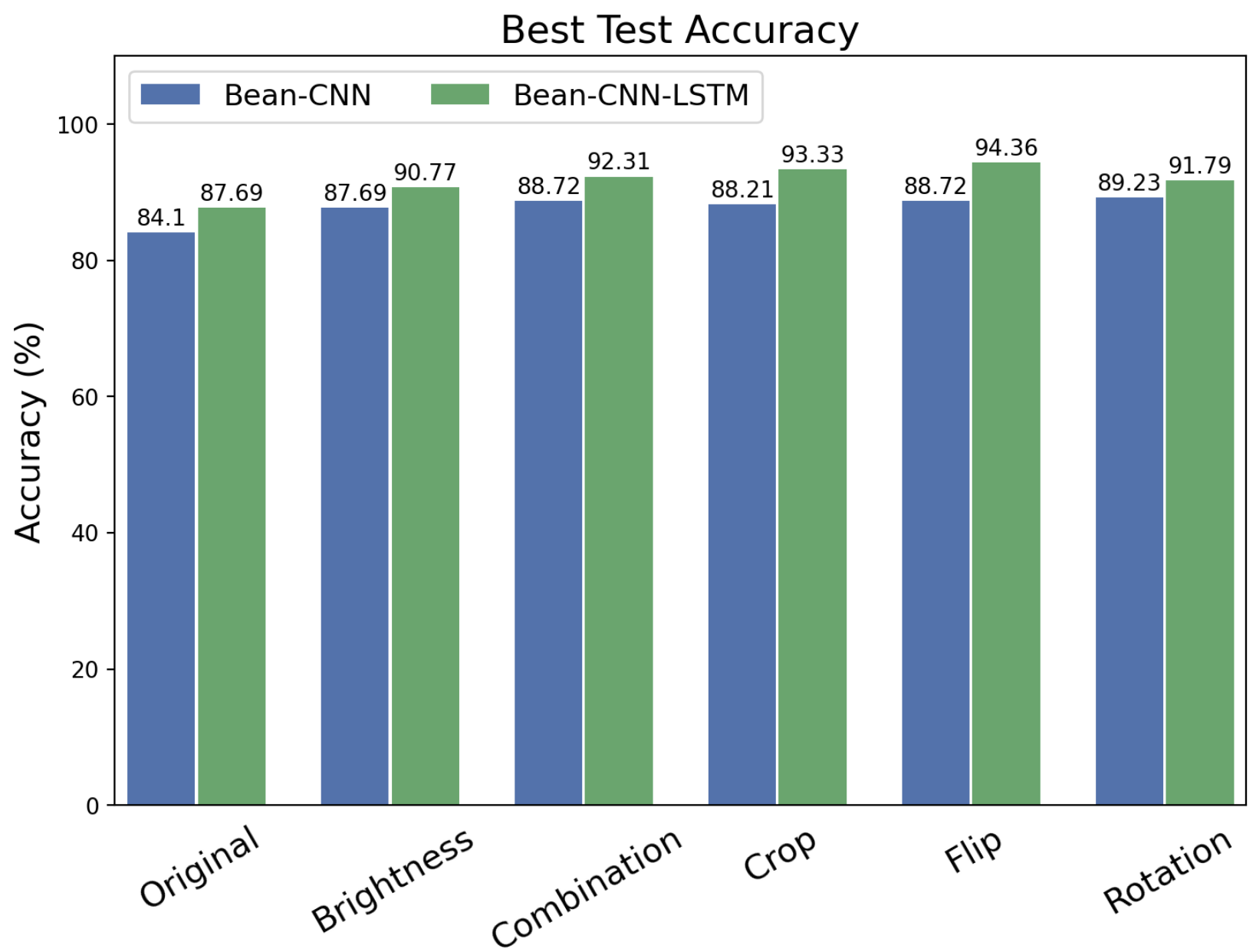}
    \caption{Graph representation of the best test accuracy: Comparison of two custom lightweight architectures}
    \label{fig:fig7}
\end{figure}

\begin{table}
\caption{\textbf{Top-1 Test performance of custom lightweight models}: {Results across training sets - the best shown in bold }}
\setlength{\tabcolsep}{3pt}

\begin{tabular}{|p{60pt}|p{35pt}|p{35pt}|p{35pt}|p{35pt}|}
\hline
\multicolumn{5}{|c|}{\textbf{Bean-CNN}} \\
\hline
Training set &
Accuracy &
Loss &
F1 Score &
MCC \\
\hline

Original & 0.8410 & 0.5180 & 0.8410 & 0.7614 \\
Brightness & 0.8769 & 0.5180 & 0.8751 & 0.8163 \\
Combination & 0.8872 & 0.6561 & 0.8882 & 0.8329 \\
Crop & 0.8821 & 0.4597 & 0.8844 & 0.8267 \\
Flip & 0.8872 & 0.5089 & 0.8867 & 0.8307 \\
Rotation & \textbf{0.8923} & 0.5130 & \textbf{0.8913} & \textbf{0.8391} \\
\hline

\hline
\multicolumn{5}{|c|}{\textbf{Bean-CNN-LSTM}} \\
\hline
Training set &
Accuracy &
Loss &
F1 Score &
MCC \\
\hline
Original & 0.8769 & 0.3773 & 0.8737 & 0.8173 \\
Brightness & 0.9077 & 0.3902 & 0.9074 & 0.8615 \\
Combination & 0.9231 & 0.4116 & 0.9234 & 0.8856 \\
Crop & 0.9333 & 0.2757 & 0.9332 & 0.9000 \\
Flip & \textbf{0.9436} & 0.1769 & \textbf{0.9438} & \textbf{0.9164} \\
Rotation & 0.9179 & 0.3398 & 0.9168 & 0.8776 \\
\hline
\end{tabular}
\label{tab:tab2}
\end{table}

Figures \ref{fig:fig5} and \ref{fig:fig6} visualise the training progress of the best models trained on the original training set (905 samples). Table \ref{tab:tab2} describes the best performance across all training sets (the original and the augmented), which is summarised in Figure \ref{fig:fig7}. Bean-CNN-LSTM consistently outperforms Bean-CNN, and the accuracy varies by 2.56\% to 5.64\%, depending on the training set. A smaller disparity between accuracy and MCC in Bean-CNN-LSTM implies that this model made fewer false predictions than the Bean-CNN models. 

\begin{figure}
    \centering
    \includegraphics[width=0.93\linewidth]{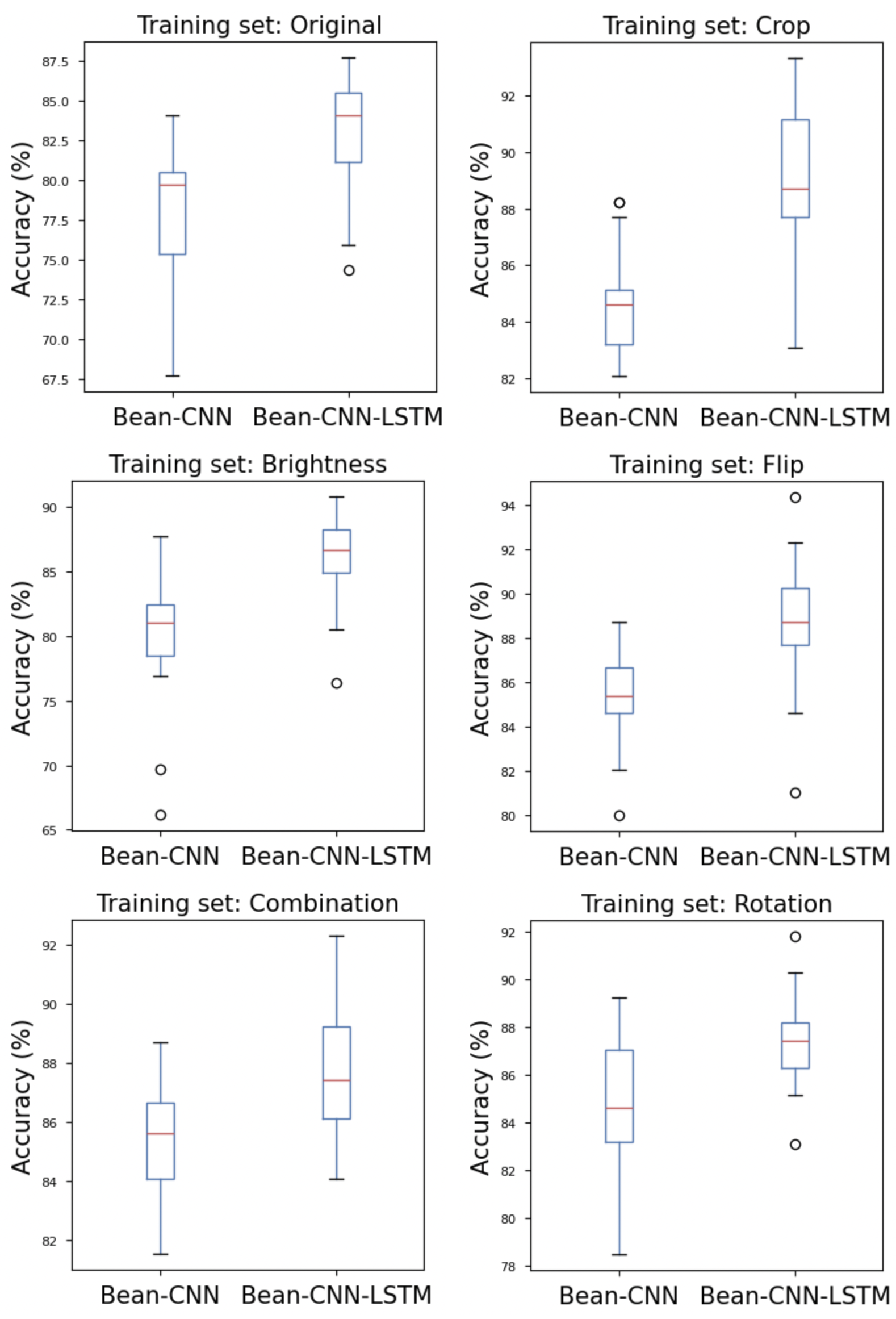}
    \caption{The average test accuracy on each training set: The red line indicates the median value, and a box represents the distribution between 25 and 75 percentiles. Small hollow circles indicate outliers.}
    \label{fig:fig8}
\end{figure}

We also analyse the average results from all 30 training runs for a more general tendency. Figure \ref{fig:fig8} represents these results in the box and whisker plots, showing the central distribution. The overall accuracy metrics correspond to the best performance, implying that the hybrid CNN-LSTM is more effective than the baseline CNN, with noticeable gaps.

\subsection{Results: Effect of augmentation to custom lightweight models}

\begin{figure}
    \centering
    \includegraphics[width=0.65\linewidth]{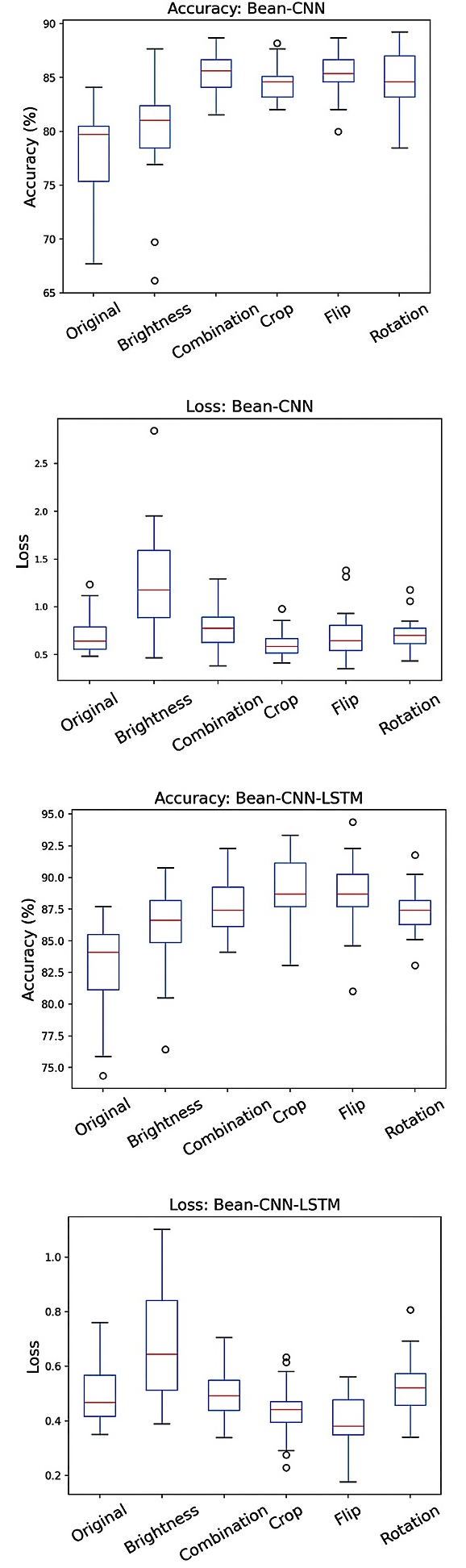}
    \caption{Box-plot representation: The average performance of our lightweight custom models (Bean-CNN and Bean-CNN-LSTM) from 30 training runs}
    \label{fig:fig9}
\end{figure}

Training on the augmented sets, as shown in Table \ref{tab:tab2},  generally increases the initial accuracy to a different extent. Bean-CNN shows the highest accuracy on the rotation set, increasing the initial accuracy by 5.13\%. The best Bean-CNN-LSTM improves accuracy by 6.67\% on the flip set. Both models on the brightness set show the least increased accuracy with elevated loss. The multi-technique combination also raises accuracy and loss simultaneously, indicating a greater uncertainty in their performance\cite{goodfellow_deep_2016}.
The average performance results of our custom models also show notable trends. As shown in Figure \ref{fig:fig9}, the crop, flip, rotation, and combination methods equally improve Bean-CNN. The loss improves further with the crop technique, indicating an additional benefit to the performance. For Bean-CNN-LSTM, both the flip and the crop techniques are more effective than others when all metrics are considered. Data analysis of the training sets does not show strong correlations between each training set and performance, although we found that the brightness set has elevated frequency and noise levels, which may have negatively contributed to the performance of all models. 

From a data governance perspective, these results highlight the necessity of principled data curation over exhaustive data expansion. Our findings indicate that while generic augmentation combinations are often viewed as a default for increasing model robustness, they can introduce a form of synthetic noise that degrades the internal consistency of lightweight models. For agricultural datasets, data governance must prioritise transformations that preserve the biological integrity of the specimen. Geometric shifts like flipping and cropping mimic natural variations in camera angles while maintaining the distinctive morphological features of the lesions. In contrast, excessive photometric noise can obscure the subtle colour gradients critical for distinguishing early-stage rust from nutrient deficiencies. Thus, managing the quality of synthetic data through tailored augmentation is as critical to system reliability as the volume of the underlying training set.

\subsection{Results: The best-performing custom lightweight model}
\begin{figure}
    \centering
    \includegraphics[width=0.7\linewidth]{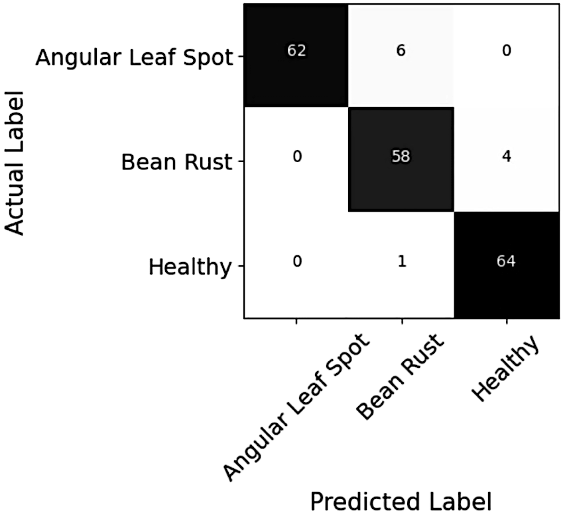}
    \caption{Confusion matrix: The best Bean-CNN-LSTM model }
    \label{fig:fig10}
\end{figure}

The best-performing custom model was the Bean-CNN-LSTM trained on the flip training set. It showed an accuracy of 94.36\%, a weighted average F1 score of 94.38\%, and an MCC of 91.64\%. Table \ref{tab:tab3} describes its class-specific performance. The recall score indicates the most false-negative cases in the angular leaf spot class. The precision score shows that the bean rust class has the most false-positive cases. The F1 scores suggest that the most balanced performance is in the healthy class, whereas the least is in the bean rust class. The confusion matrix in Figure \ref{fig:fig10} visualises this performance and shows a higher error pattern in the bean rust class. 

\begin{table}
\caption{\textbf{Performance of the best Bean-CNN-LSTM}}
\label{table}
\setlength{\tabcolsep}{3pt}
\begin{tabular}{|p{80pt}|p{35pt}|p{30pt}|p{35pt}|}
\hline
Class &
Precision (\%)&
Recall (\%)&
F1 Score (\%)\\
\hline
Angular leaf spot & 100.00 & 91.18 & 95.38\\
Bean rust & 89.23  & 93.55 & 91.34 \\
Healthy & 94.12 & 98.46 & 96.24 \\
\hline\hline
Overall Acc.(\%) & \multicolumn{3}{|c|}{94.36} \\
Overall F1(\%) & \multicolumn{3}{|c|}{94.38} \\
\hline
\end{tabular}
\label{tab:tab3}
\end{table}

\subsection{Results: Comparison with existing models}
To investigate the robustness of our CNN-LSTM-based method, we conducted further experiments with a pre-trained deep network, EfficientNet-B7 \cite{tan_efficientnet_2019}, trained on a larger version of the \textit{iBean} dataset. We selected the EfficientNet for its excellent performance in image recognition, yet improved efficiency compared to other CNN-based models. The B7 model is even smaller/faster and achieves better image recognition accuracy.

The \textit{ibean} dataset was augmented to create a 40 times larger training and validation set by using geometric and photometric transformations, including random rotation, horizontal and vertical flips, scaling, blurring and contrast/brightness adjustment. Much like the original dataset, the resulting augmented dataset is nearly balanced across the classes, containing a total of 41,300 training images - 13,800, 13,900 and 13,600 images for angular leaf spot, bean rust and healthy leaves, respectively. It also contains 5,320 validation images - 1760, 1800 and 1760 images for angular leaf spot, bean rust and healthy leaves, respectively. To allow for a fair comparison across previous models, the test set remains the 128 images provided in the original dataset. 

To allow for some comparison between conventional Dense layers and LSTM layers, we created two separate models from EfficientNet by replacing its final classification layer. In the first model, we replaced the final classification layer with Dense or fully-connected (FC) layers, interspersed with Dropout (30\% - 50\%) and batch normalisation layers. In the second model, the classification layer is replaced with an LSTM layer and a linear classification layer, and a Dropout layer (50\%) in between them. We refer to the former model as the EfficientNetB7+FC model and the latter as the EfficientNetB7+LSTM model. For best results, the entire architecture was trained, but the EfficientNet backbone was unfrozen in blocks after every 5 epochs. Both models were trained with the Adam optimiser at a learning rate of 5e-4, cross-entropy loss function, a batch size of 8 and an image size of 380 $\times$ 380. The Adam optimiser has proven to be more efficient for image recognition/classification tasks than most other variants of the gradient descent algorithm, such as Stochastic Gradient Descent (SGD) and RMSProp. This is due to Adam's adaptive learning rate and momentum-based weight updates, which lead to faster convergence \cite{kingma2014adam}. The very low learning rate slowed down the training but allowed for a smoother training with more chances of finding the global minima and better generalisation. The batch size of 8 was necessary to allow for more memory management and better generalisation while training a very deep network like EfficientNet, but it also slows down training. We chose to increase the image resolution performance a bit more here to allow for improved performance, i.e. more pixels for convolution and sequencing.

The resulting performances of the models are not unexpected. Table \ref{tab:tab4} shows the results of both models on the test set; they both reached overall accuracies and F1 scores of 99.22\% and had the same scores (precision, recall and F1) across the classes. Both models reached training and validation accuracies of about 98\%, but the EfficientNetB7+FC model reached this performance in only 3 epochs, while the EfficientNetB7+LSTM model reached it in 16 epochs. This means that the FC-based model did not even require the first block of layers of the EfficientNet backbone to be unfrozen before it reached such an excellent classification performance, while the LSTM layer needed some blocks of layers of the backbone to be unfrozen. This shows that the linear layers in the FC-based model are more effective in leveraging the features from the EfficientNet backbone to achieve both high accuracy and efficiency in the bean leaf disease classification task, while the fact that the LSTM-based model achieved the same result with more training indicates that it could not efficiently leverage the huge image features from the EfficientNet backbone for image classification. In fact, to simulate more similarity with the FC-based model, we also included batch normalisation and adaptive average pooling in the LSTM-based model, but the result was rather degraded, reaching only 97.68\% test F1 score after 16 epochs. This could also be due to the combined augmentation techniques used in the extended dataset, as our earlier results indicate that the combination of several augmentation techniques did not necessarily produce the best result for both models. However, the fact that the EfficientNetB7+LSTM model still reaches the same accuracy as the EfficientNetB7+FC model after more training epochs indicates that it is suitable for the bean classification task and indeed smaller in size, as shown in Table \ref{tab:tab6}.

\begin{table}
\caption{\textbf{Classification scores of both the EfficientNetB7+FC \& EfficientNetB7+LSTM Models}}
\label{table}
\setlength{\tabcolsep}{3pt}
\begin{tabular}{|p{80pt}|p{35pt}|p{30pt}|p{35pt}|}
\hline
Class &
Precision (\%) &
Recall (\%) &
F1 score (\%)\\
\hline
Angular leaf spot & 97.73 & 100.00 & 98.85\\
Bean rust & 100.00 & 97.67 & 98.82\\
Healthy & 100.00 & 100.00 & 100.00\\
\hline\hline
Overall Acc.(\%) & \multicolumn{3}{|c|}{99.22} \\
Overall F1(\%) & \multicolumn{3}{|c|}{99.22} \\
\hline
\end{tabular}
\label{tab:tab4}
\end{table}

\begin{table}
\caption{\textbf{Comparison with previous works}}
\label{table}
\setlength{\tabcolsep}{3pt}
\begin{tabular}{|p{75pt}|p{80pt}|p{25pt}|p{25pt}|}
\hline
Source &
Method &
Acc. (\%) &
F1 Score (\%)\\
\hline
Abed et al., 2021 \cite{abed_modern_2021} & DenseNet121 & 91.02 & -- \\
Elfatimi et al., 2022 \cite{elfatimi_beans_2022} & MobileNetV2 & 92.97 & 92.94 \\
Singh et al., 2023 \cite{singh_classification_2023} & EfficientnetB6 & 91.74 & -- \\
{\"O}nler 2023 \cite{onler_feature_2023} & MobileNetV2 & 99.24 & -- \\
Suntoyo et al., 2024 \cite{sunyoto_innovative_2024} & DenseNet121 & 96.90 & 97.00 \\
Jain \& Aneja 2025 \cite{Jain_automated_2025} & InceptionV3 & 91.00 & 91.00 \\
Karthik et al., 2025 \cite{Karthik_explainable_2025} & Transformer+CNN & 97.66 & 97.67 \\
\hline
\textbf{Ours} & \textbf{Bean-CNN-LSTM} & \textbf{94.36} & \textbf{94.38} \\
\textbf{} & \textbf{EfficientNetB7 + FC} & \textbf{99.22} & \textbf{99.22} \\
\textbf{} & \textbf{EfficientNetB7 + LSTM} & \textbf{99.22} & \textbf{99.22} \\
\hline
\end{tabular}
\label{tab:tab5}
\end{table}

Table \ref{tab:tab5} shows a comparison of the performance of previous models on the \textit{iBean} dataset, with the last three rows referring to our models. It is important to state that some of the high-scoring models, such as \cite{onler_feature_2023}, reported only the classification accuracy, which is not a guarantee of good generalisation. Although the F1 score of each class was reported up to two decimal places (1.00, 0.99, 0.99), the overall F1 score across all 3 classes was not reported. The reader is therefore left to take an average of the class-based F1 scores (99.3\%), and this computation often leads to a higher score, especially because the floating point precision is even lower (up to 2 decimal places). If we also averaged the class-based F1 scores from our model (Table \ref{tab:tab4}) up to 2 decimal places, we would report a higher, and in fact, the same F1 score of 99.3\%. It is also noteworthy that in many of these works, such as \cite{onler_feature_2023}, \cite{Karthik_explainable_2025}, and \cite{sunyoto_innovative_2024}, the models were trained for a high number of epochs, 100, 50, and 30, respectively. This is most likely because they (except \cite{Karthik_explainable_2025}) used the transfer learning approach of training only a few custom layers on top of the pre-trained backbone architecture. While this approach can be computationally less expensive, it does not guarantee the best performance and is prone to overfitting due to the generic image features transferred from the backbone, hence the need for longer training. Overall, both of our pre-trained models (EfficientNetB7+FC, EfficientNetB7+LSTM) achieved better generalisation in fewer training iterations than the previous models. Note that we randomly selected 195 samples as the test subset for the custom (Bean-CNN \& Bean-LSTM) models in our initial experiments, as described in the methodology section, while the successive experiments with EfficientNet used the 128 test data samples, as provided in the \textit{ibean} dataset \cite{noauthor_ai-lab-makerere_2020}. This is to facilitate a fair comparison with existing models.

\section{Discussion}
As summarised in Table \ref{tab:tab6}, adopting an LSTM has a strong advantage for our ultra-lightweight custom models. This hybrid architecture can be an optimal solution in resource-constrained hardware environments such as embedded/edge devices. Note that in Table \ref{tab:tab6}, the EfficientNetB7+LSTM model has two different values in the third column because the number of training parameters increased after more layers of the EfficientNetB7 model were unfrozen during training. However, the remarkably high accuracy despite the smaller model size/number of parameters indicates that the LSTM-based model is an effective approach.

Beyond technical performance, the deployment of this lightweight CNN-LSTM architecture offers significant advantages for integrated agricultural data management. By achieving state-of-the-art results with a model size significantly smaller than traditional CNNs (1.86 MB vs. 6.11 MB for our custom models), this system facilitates the real-time processing of leaf health data directly on edge devices. In a management context, this decentralised data processing reduces the reliance on costly cloud infrastructure and high-bandwidth connectivity, which are often absent in the rural regions where bean production is most vital. Consequently, this approach enables more agile farm management decisions, allowing for immediate localised interventions that can mitigate the estimated 47\% to 55\% yield losses typically associated with bean rust and angular leaf spot.

\begin{figure}
    \centering
    \includegraphics[width=0.8\linewidth]{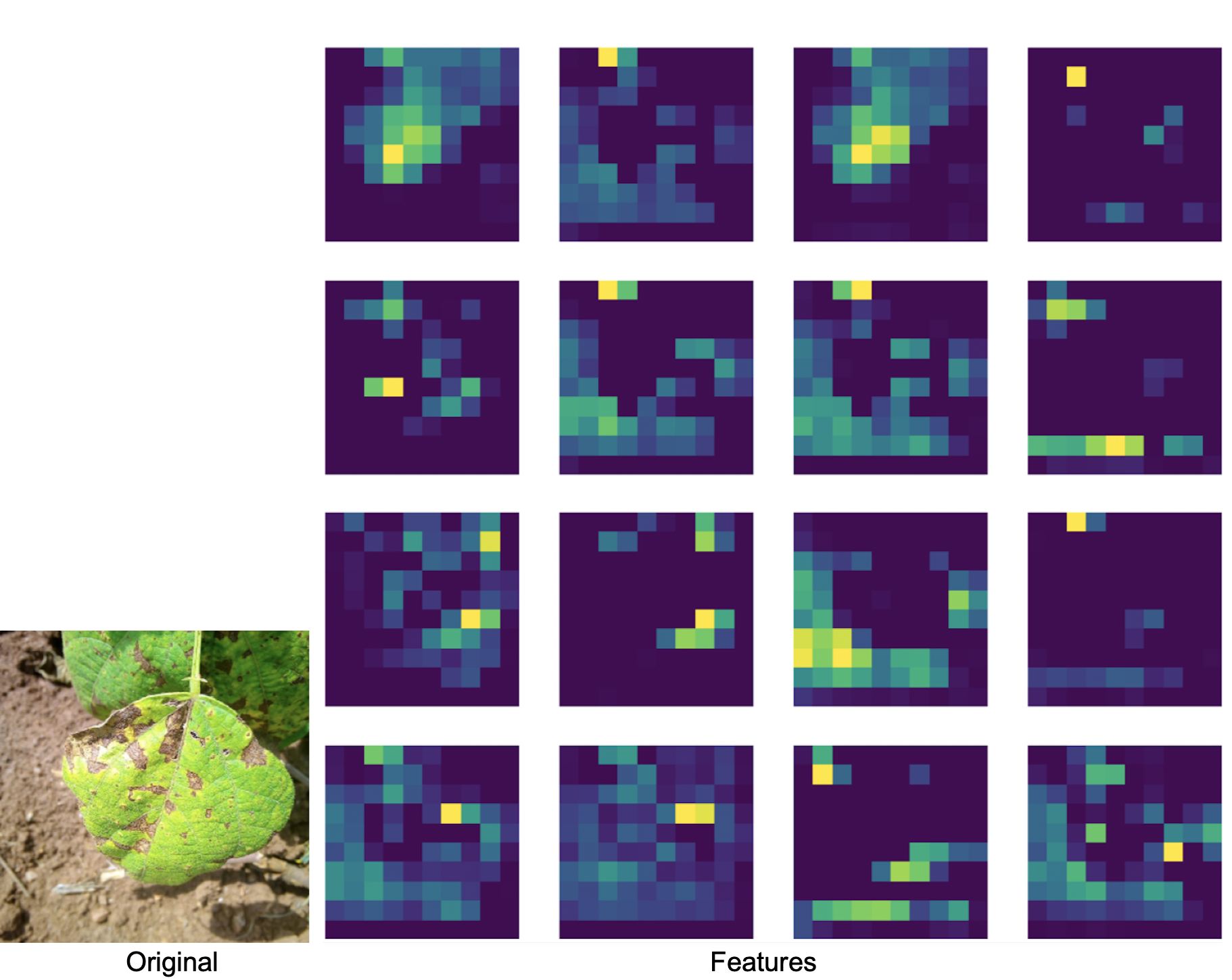}
    \caption{GradCAM Feature maps. The heatmaps indicate that the model correctly prioritises the necrotic centres of the lesions rather than the leaf edges, validating the management reliability of the system.}
    \label{fig:fig11}
\end{figure}

\begin{table}
\caption{\textbf{Our model summary}}
\label{table}
\setlength{\tabcolsep}{3pt}
\begin{tabular}{|p{60pt}|p{40pt}|p{50pt}|p{40pt}|}
\hline
Models &
Model size (MB) &
Number of parameters &
Best accuracy (\%)\\
\hline
\textbf{Bean-CNN}  & 6.11 & 527,171 & 89.23\\
\textbf{Bean-CNN-LSTM}  & 1.86 & 155,219 & 94.38 \\
\textbf{EfficientNetB7 + FC} & 260.56 & 67,825,595 & 99.22 \\
\textbf{EfficientNetB7 + LSTM} & 250.20 & 1,377,667 / 52,844,795 & 99.22 \\
\hline
\end{tabular}
\label{tab:tab6}
\end{table}
The dynamic weight computations of an LSTM clearly showed strength in discovering spatiotemporal correlations within the feature map (Figure \ref{fig:fig11}) in the ibean data. Our experiment with input vectors resulted in reduction in the number of features. Despite this, Bean-CNN-LSTM required up to 1.7 times longer training time than Bean-CNN. Nevertheless, Bean-CNN is a more complex model with probable redundant connections \cite{kahatapitiya_exploiting_2021}, which may explain the struggle of this model to achieve the same level of performance as Bean-CNN-LSTM.

\begin{figure}
    \centering
    \includegraphics[width=0.75\linewidth]{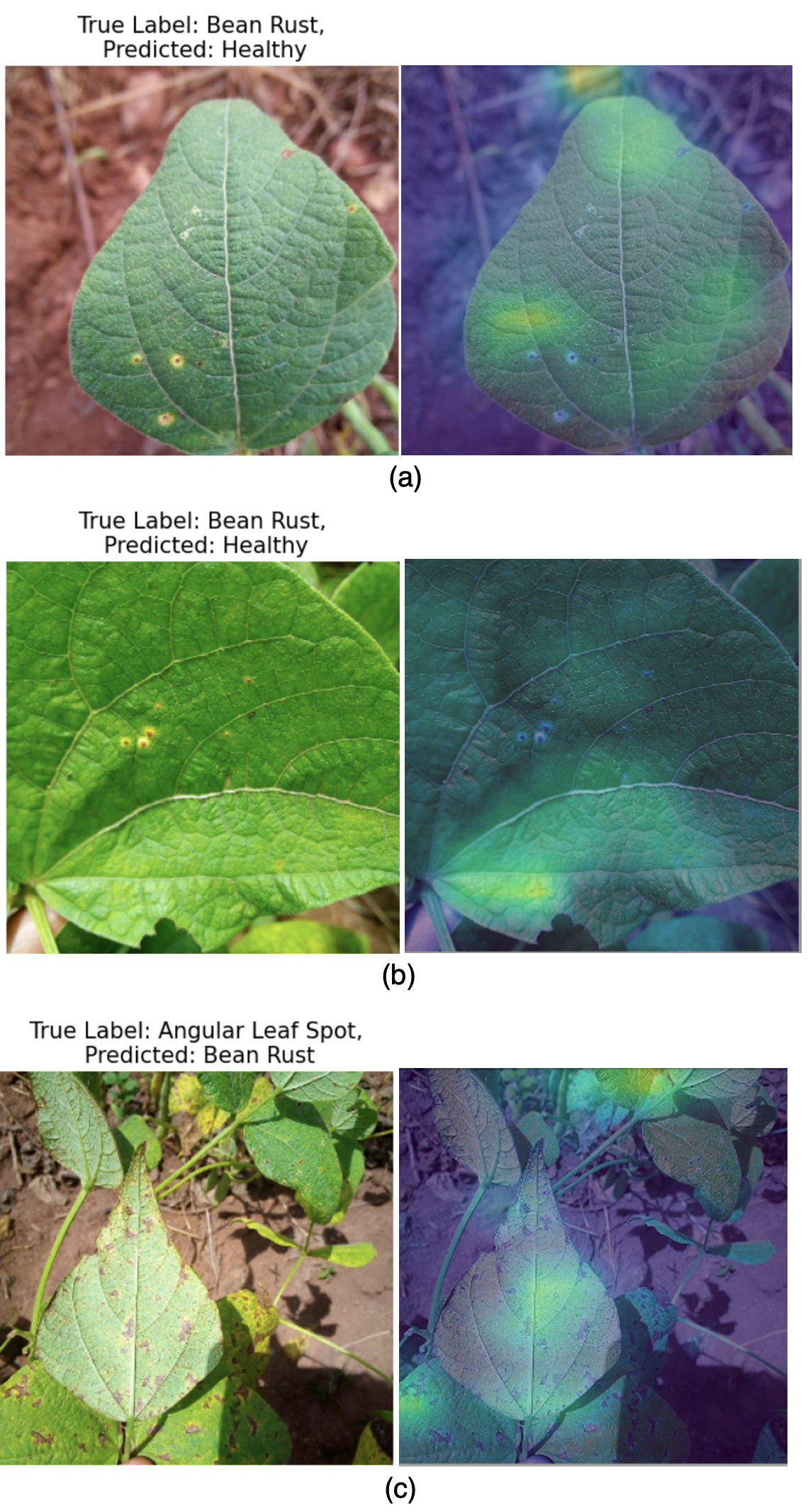}
    \caption{Misclassified images}
    \label{fig:fig12}
\end{figure}

Despite the strong performance of our compact models, we observed limitations of these models, and the superior and sophisticated performance of the pre-trained models in this task. More precisely, the best Bean-CNN-LSTM showed the weakest performance in detecting bean rust. Through visual examination of the data, we found that the bean rust symptoms are less severe and more varied, which could be more challenging for the model, explaining the higher error rates. \cite{barbedo_review_2016}. Figure \ref{fig:fig12} shows the misidentification by this model. Similar to Figure \ref{fig:fig12}a, Figure \ref{fig:fig12}b represents the early onset of infection with additional difficulty in a highly close-up image. Figure \ref{fig:fig12}c represents an error in the image with a busy background. We consider that a busy background in training data can improve generalisability in real-life conditions, as previously attempted \cite{sun_mean-ssd_2021, fenu_diamos_2021}. Thus, there is still room for improvement in our ultra-lightweight model in this regard. 


In addition, our systematic review of augmentation techniques suggests that generic combinations may introduce counterproductive noise for specialised agricultural tasks. Specifically, while standard multi-technique pipelines are often used as a default, our results indicate that they can simultaneously raise both accuracy and loss, signalling increased model uncertainty. We posit that for leaf pathology, geometric transformations, such as flipping and cropping, preserve the essential structural features of disease lesions, such as the distinctive brown angular patterns of leaf spot, whereas excessive photometric shifts (like extreme brightness or random noise) may obscure the subtle halo effects critical for identifying early-stage rust. This underscores the necessity of tailored augmentation strategies that prioritise the preservation of domain-specific visual markers over simple dataset expansion.

\subsection{Implications for Agricultural Resource Management}
The transition from centralised high-performance computing to decentralised Edge-AI carries significant implications for agricultural resource management. In traditional management models, the latency and cost of data transmission to the cloud often result in delayed interventions. By reducing the model footprint to 1.86 MB, a 70\% reduction compared to standard architectures, this research facilitates the integration of sophisticated diagnostics into low-cost mobile management systems. This efficiency allows for Precision Agriculture practices where farmers can manage specific sections of a plot based on real-time data, optimising the application of fungicides and labour. Such data-driven decision-making not only reduces the operational overhead for smallholder farmers but also mitigates the environmental impact of broad-spectrum chemical use, aligning agricultural productivity with sustainable management goals.

\subsection{Limitations}
This work has a number of limitations, and caution is necessary not to overgeneralise the findings. The results are subject to the architectures, and the ML settings described in this article for experimental convenience. Despite the remarkable performance of the models, the tuning and training process can be time-consuming. Although care was taken to avoid the models' overfitting, the significantly small training data necessitates more diverse and robust sets to ensure that the proposed model has adequate generalisation. Furthermore, the successful model construction of this work does not directly indicate real-time performance across crop types since the training, validation and testing are based solely on the common bean plant cultivated in Uganda. Other crop types or bean plant varieties may exhibit different results.

\section{Conclusion}
Automated image-based disease classification aims to reduce yield loss and ensure global food security. We investigated the feasibility of a hybrid CNN-LSTM architecture for this task. In our two-phased experiments, we introduced LSTM-based ultra-lightweight custom models and pre-trained EfficientNet. Our primary results show that Bean-CNN-LSTM consistently outperforms Bean-CNN with 2.56\% to 5.64\% higher best test accuracy, depending on training sets. The best Bean-CNN-LSTM model of 1.9 MB in size achieved an accuracy of 94.36\%. This result inspired us to conduct further experiments with a pre-trained image recognition model, EfficientNet. In these experiments, we discovered that our modified EfficientNet models, trained on significantly larger data, improved the performance further. Both an NN-based and an LSTM-based model achieved 99.22\% accuracy and F1 score. Although there are no significant gains in training efficiency, integrating an LSTM with CNN contributed to streamlining the model and enhancing performance. Nevertheless, we found that complex backgrounds and disease pattern variations are challenging, especially for our lightweight models. For further improvement, advanced feature extraction methods using specialised filters or segmentation are worth exploring. Also, comparing other RNN families, such as a gated recurrent unit (GRU), could shed light on discovering more efficient models.

This study contributes to the field of data science management by demonstrating that model complexity is not a prerequisite for high performance. The proposed 1.86 MB CNN-LSTM framework offers a scalable solution for integrating AI into agricultural management systems, bridging the gap between high-level data analytics and practical, on-field decision support.

\bibliographystyle{elsarticle-num}
\bibliography{references}

@article{pamela_severity_2014,
	title = {Severity of angular leaf spot and rust diseases on common beans in {Central} {Uganda}},
	volume = {14},
	url = {https://www.ajol.info/index.php/ujas/article/view/126189},
	number = {1},
	journal = {Uganda Journal of Agricultural Sciences},
	author = {Pamela, P. and Mawejje, D. and Ugen, M. },
	year = {2014},
}

@article{venbrux_current_2023,
	title = {Current and emerging trends in techniques for plant pathogen detection},
	volume = {14},
	url = {https://doi.org/10.3389/fpls.2023.1120968},
	doi = {10.3389/fpls.2023.1120968},
	journal = {Frontiers in Plant Science},
	author = {Venbrux, M. and Crauwels, S. and Rediers, H.},
	year = {2023},
}

@article{mahlein_plant_2016,
	title = {Plant disease detection by imaging sensors–parallels and specific demands for precision agriculture and plant phenotyping},
	volume = {100},
	url = {https://doi.org/10.1094/PDIS-03-15-0340-FE},
	doi = {10.1094/PDIS-03-15-0340-FE},
	number = {2},
	journal = {Plant Disease},
	author = {Mahlein, A.-K.},
	year = {2016},
	pages = {241--251},
}

@article{slimani_artificial_2023,
	title = {Artificial {Intelligence}-based {Detection} of {Fava} {Bean} {Rust} {Disease} in {Agricultural} {Settings}: {An} {Innovative} {Approach}},
	volume = {14},
	doi = {doi.org/10.14569/IJACS},
	number = {6},
	journal = {International Journal of Advanced Computer Science \& Applications},
	author = {Slimani, H},
	year = {2023},
	pages = {119--128},
}

@article{mohanty_using_2016,
	title = {Using {Deep} {Learning} for {Image}-{Based} {Plant} {Disease} {Detection}},
	volume = {7},
	doi = {doi.org/10.3389/fpls.2016.01419},
	journal = {Frontiers in Plant Science},
	author = {Mohanty, S. P.},
	year = {2016},
	pages = {1419},
}

@article{islam_combined_2020,
	title = {A combined deep {CNN}-{LSTM} network for the detection of novel coronavirus ({COVID}-19) using {X}-ray images},
	volume = {20},
	doi = {doi.org/10.1016/j.imu.2020.100412},
	journal = {Informatics in Medicine Unlocked},
	author = {Islam, Z. and Islam, M. and Amanullah, Asraf},
	year = {2020},
	pages = {100412--100412},
}

@article{donahue_long-term_2014,
  author={Donahue, Jeff and Hendricks, Lisa Anne and Rohrbach, Marcus and Venugopalan, Subhashini and Guadarrama, Sergio and Saenko, Kate and Darrell, Trevor},
  journal={IEEE Transactions on Pattern Analysis and Machine Intelligence}, 
  title={Long-Term Recurrent Convolutional Networks for Visual Recognition and Description}, 
  year={2017},
  volume={39},
  number={4},
  pages={677-691},
  doi={10.1109/TPAMI.2016.2599174}
}

@inproceedings{jian_support_2010,
	title = {Support vector machine for recognition of cucumber leaf diseases},
	volume = {5},
	doi = {10.1109/ICACC.2010.5487242},
	booktitle = {2010 2nd {International} {Conference} on {Advanced} {Computer} {Control}},
	author = {Jian, Zhang and Wei, Zhang},
	year = {2010},
	pages = {264--266},
}

@article{lu_identification_2017,
	title = {Identification of rice diseases using deep convolutional neural networks},
	volume = {267},
	issn = {0925-2312},
	url = {https://www.sciencedirect.com/science/article/pii/S0925231217311384},
	doi = {https://doi.org/10.1016/j.neucom.2017.06.023},
	journal = {Neurocomputing},
	author = {Lu, Yang and Yi, Shujuan and Zeng, Nianyin and Liu, Yurong and Zhang, Yong},
	year = {2017},
	pages = {378--384},
}

@article{g_identification_2019,
	title = {Identification of plant leaf diseases using a nine-layer deep convolutional neural network},
	volume = {76},
	issn = {0045-7906},
	url = {https://www.sciencedirect.com/science/article/pii/S0045790619300023},
	doi = {https://doi.org/10.1016/j.compeleceng.2019.04.011},
	journal = {Computers \& Electrical Engineering},
	author = {G. Geetharamani and J. {Arun Pandian}},
	year = {2019},
	pages = {323--338},
}

@article{patil_plant_2023,
	title = {Plant Leaf Disease Classification Using Optimal Tuned Hybrid LSTM-CNN Model.},
	volume = {4},
	doi = {doi.org/10.1007/s42979-023-02245-7},
	number = {6},
	journal = {SN Computer Science},
	author = {Patil, M. A. and Manohar, M.},
	year = {2023},
	pages = {710},
}

@inproceedings{devi_plant_2023,
	title = {Plant Disease Classification using CNN-LSTM Techniques},
	doi = {10.1109/ICSSIT55814.2023.10061003},
	booktitle = {2023 5th {International} {Conference} on {Smart} {Systems} and {Inventive} {Technology} ({ICSSIT})},
	author = {Devi, E.Anna and Gopi, S. and Padmavathi, U. and Arumugam, Sajeev Ram and Premnath, S.P. and Muralitharan, Divya},
	year = {2023},
	pages = {1225--1229},
}

@article{abade_plant_2021,
	title = {Plant diseases recognition on images using convolutional neural networks: {A} systematic review},
	volume = {185},
	issn = {0168-1699},
	url = {https://www.sciencedirect.com/science/article/pii/S0168169921001435},
	doi = {https://doi.org/10.1016/j.compag.2021.106125},
	journal = {Computers and Electronics in Agriculture},
	author = {André Abade and Paulo Afonso Ferreira and Flavio {de Barros Vidal}},
	year = {2021},
	pages = {106125},
}

@article{elfatimi_beans_2022,
	title = {Beans {Leaf} {Diseases} {Classification} {Using} {MobileNet} {Models}},
	volume = {10},
	doi = {10.1109/ACCESS.2022.3142817},
	journal = {IEEE Access},
	author = {Elfatimi, Elhoucine and Eryigit, Recep and Elfatimi, Lahcen},
	year = {2022},
	pages = {9471--9482},
}

@inproceedings{singla_deep_2024,
	title = {Deep {Learning} based {Bean} {Leaf} {Lesion} {Classification} utilizing {EfficientNetV2}-{S}},
	doi = {10.1109/I-SMAC61858.2024.10714612},
	booktitle = {2024 8th {International} {Conference} on {I}-{SMAC} ({IoT} in {Social}, {Mobile}, {Analytics} and {Cloud}) ({I}-{SMAC})},
	author = {Singla, Seerat and Gupta, Rupesh},
	year = {2024},
	pages = {1394--1399},
}

@article{rodriguez-lira_comparative_2024,
	title = {Comparative {Analysis} of {YOLO} {Models} for {Bean} {Leaf} {Disease} {Detection} in {Natural} {Environments}},
	volume = {6},
	issn = {2624-7402},
	url = {https://www.mdpi.com/2624-7402/6/4/262},
	doi = {10.3390/agriengineering6040262},
	number = {4},
	journal = {AgriEngineering},
	author = {Rodríguez-Lira, Diana-Carmen and Córdova-Esparza, Diana-Margarita and Álvarez-Alvarado, José M. and Romero-González, Julio-Alejandro and Terven, Juan and Rodríguez-Reséndiz, Juvenal},
	year = {2024},
	pages = {4585--4603},
}

@inproceedings{hohman_model_2024,
	series = {{CHI} ’24},
	title = {Model {Compression} in {Practice}: {Lessons} {Learned} from {Practitioners} {Creating} {On}-device {Machine} {Learning} {Experiences}},
	url = {http://dx.doi.org/10.1145/3613904.3642109},
	doi = {10.1145/3613904.3642109},
	booktitle = {Proceedings of the {CHI} {Conference} on {Human} {Factors} in {Computing} {Systems}},
	publisher = {ACM},
	author = {Hohman, Fred and Kery, Mary Beth and Ren, Donghao and Moritz, Dominik},
	month = may,
	year = {2024},
	pages = {1--18},
}

@article{sun_mean-ssd_2021,
	title = {{MEAN}-{SSD}: {A} novel real-time detector for apple leaf diseases using improved light-weight convolutional neural networks},
	volume = {189},
	issn = {0168-1699},
	url = {https://www.sciencedirect.com/science/article/pii/S0168169921003963},
	doi = {https://doi.org/10.1016/j.compag.2021.106379},
	journal = {Computers and Electronics in Agriculture},
	author = {Sun, Henan and Xu, Haowei and Liu, Bin and He, Dongjian and He, Jinrong and Zhang, Haixi and Geng, Nan},
	year = {2021},
	pages = {106379},
}

@article{yang_image_2023,
	title = {Image {Data} {Augmentation} for {Deep} {Learning}: {A} {Survey}},
    journal = {},
	howpublished = {https://arxiv.org/abs/2204.08610},
	author = {Yang, Suorong and Xiao, Weikang and Zhang, Mengchen and Guo, Suhan and Zhao, Jian and Shen, Furao},
	year = {2023},
	note = {arXiv-eprint: 2204.08610},
}

@misc{noauthor_ai-lab-makerere_2020,
	title = {{AI}-{Lab}-{Makerere} / ibean},
	howpublished = {https://github.com/AI-Lab-Makerere/ibean/},
	publisher = {Makerere AI lab},
        author = {},
	month = jan,
	year = {2020},
}

@book{a_muimba-kankolongo_food_2018,
	title = {Food {Crop} {Production} by {Smallholder} {Farmers} in {Southern} {Africa}},
	url = {https://www.sciencedirect.com/book/9780128143834/food-crop-production-by-smallholder-farmers-in-southern-africa},
	publisher = {Science Direct},
	author = {{A.Muimba-Kankolongo}},
	year = {2018},
}

@article{deng_ensemble_2014,
	title = {Ensemble deep learnig for speech recognition},
	volume = {1},
	doi = {doi:10.21437/Interspeech.2014-433},
	journal = {Interspeech},
	author = {Deng, L. and Platt, J. C.},
	year = {2014},
}

@inproceedings{sainath_convolutional_2015,
	title = {Convolutional, {Long} {Short}-{Term} {Memory}, fully connected {Deep} {Neural} {Networks}},
	doi = {10.1109/ICASSP.2015.7178838},
	booktitle = {2015 {IEEE} {International} {Conference} on {Acoustics}, {Speech} and {Signal} {Processing} ({ICASSP})},
	author = {Sainath, Tara N. and Vinyals, Oriol and Senior, Andrew and Sak, Haşim},
	year = {2015},
	pages = {4580--4584},
}

@article{ercolano_combining_nodate,
	title = {Combining {CNN} and {LSTM} for activity of daily living recognition with a {3D} matrix skeleton representation},
	volume = {14},
	doi = {doi.org/10.1007/s11370-021-00358-7},
	journal = {Intelligent Service Robotics},
	author = {Ercolano, G. and Rossi, S.},
    year={2021},
	pages = {175--185},
}

@inproceedings{simonyan_very_2015,
  author       = {Karen Simonyan and
                  Andrew Zisserman},
  editor       = {Yoshua Bengio and
                  Yann LeCun},
  title        = {Very Deep Convolutional Networks for Large-Scale Image Recognition},
  booktitle    = {3rd International Conference on Learning Representations, {ICLR} 2015,
                  San Diego, CA, USA, May 7-9, 2015, Conference Track Proceedings},
  year         = {2015},
  url          = {http://arxiv.org/abs/1409.1556},
  bibsource    = {dblp computer science bibliography, https://dblp.org}
}

@article{zhang_comparative_2024,
  title={Comparative analysis of VGG, ResNet, and GoogLeNet architectures evaluating performance, computational efficiency, and convergence rates},
  author={Zhang, Xiao and Han, Ningning and Zhang, Jiaming},
  journal={Applied and Computational Engineering},
  volume={44},
  pages={172--181},
  year={2024}
}

@book{nixon_feature_2020,
	address = {London},
	title = {Feature {Extraction} and {Image} {Processing} for {Computer} {Vision},},
	publisher = {Academic Press},
	author = {Nixon, M. S. and Aguado, A. A.},
	year = {2020},
}

@inproceedings{he_delving_2015,
	title = {Delving {Deep} into {Rectifiers}: {Surpassing} {Human}-{Level} {Performance} on {ImageNet} {Classification}},
	doi = {10.1109/ICCV.2015.123},
	booktitle = {2015 {IEEE} {International} {Conference} on {Computer} {Vision} ({ICCV})},
	author = {He, Kaiming and Zhang, Xiangyu and Ren, Shaoqing and Sun, Jian},
	year = {2015},
	pages = {1026--1034},
}

@article{tharwat_classification_2021,
	title = {Classification assessment methods},
	volume = {17},
	doi = {doi.org/10.1016/j.aci.2018.08.003},
	number = {1},
	journal = {Applied Computing \& Informatics},
	author = {Tharwat, Alaa},
	year = {2021},
	pages = {168--192},
}

@article{chicco_advantages_2020,
	title = {The advantages of the {Matthews} correlation coefficient ({MCC}) over {F1} score and accuracy in binary classification evaluation.},
	volume = {21},
	doi = {doi: 10.1186/s12864-019-6413-7},
	number = {1},
	journal = {BMC Genomics},
	author = {Chicco, D. and Jurman, G.},
	month = jan,
	year = {2020},
}

@inproceedings{selvaraju_grad-cam_2017,
	title = {Grad-{CAM}: {Visual} {Explanations} from {Deep} {Networks} via {Gradient}-{Based} {Localization}},
	doi = {10.1109/ICCV.2017.74},
	booktitle = {2017 {IEEE} {International} {Conference} on {Computer} {Vision} ({ICCV})},
	author = {Selvaraju, Ramprasaath R. and Cogswell, Michael and Das, Abhishek and Vedantam, Ramakrishna and Parikh, Devi and Batra, Dhruv},
	year = {2017},
	pages = {618--626},
}

@book{goodfellow_deep_2016,
	title = {Deep {Learning}},
	publisher = {MIT Press},
	author = {Goodfellow, Ian and Bengio, Yoshua and Courville, Aaron},
	year = {2016},
}

@inproceedings{kahatapitiya_exploiting_2021,
	title = {Exploiting the {Redundancy} in {Convolutional} {Filters} for {Parameter} {Reduction}},
	doi = {10.1109/WACV48630.2021.00145},
	booktitle = {2021 {IEEE} {Winter} {Conference} on {Applications} of {Computer} {Vision} ({WACV})},
	author = {Kahatapitiya, Kumara and Rodrigo, Ranga},
	year = {2021},
	pages = {1409--1419},
}

@article{barbedo_review_2016,
	title = {A review on the main challenges in automatic plant disease identification based on visible range images},
	volume = {144},
	issn = {1537-5110},
	url = {https://www.sciencedirect.com/science/article/pii/S1537511015302476},
	doi = {https://doi.org/10.1016/j.biosystemseng.2016.01.017},
	journal = {Biosystems Engineering},
	author = {Barbedo, Jayme Garcia Arnal},
	year = {2016},
	pages = {52--60},
}

@article{fenu_diamos_2021,
	title = {{DiaMOS} {Plant}: {A} {Dataset} for {Diagnosis} and {Monitoring} {Plant} {Disease}},
	volume = {11},
	issn = {2073-4395},
	url = {https://www.mdpi.com/2073-4395/11/11/2107},
	doi = {10.3390/agronomy11112107},
	number = {11},
	journal = {Agronomy},
	author = {Fenu, Gianni and Malloci, Francesca Maridina},
	year = {2021},
}

@ARTICLE{Karthik_explainable_2025,
  author={Karthik, R. and Aswin, R. and Geetha, K. S. and Suganthi, K.},
  journal={IEEE Access}, 
  title={An Explainable Deep Learning Network With Transformer and Custom CNN for Bean Leaf Disease Classification}, 
  year={2025},
  volume={13},
  number={},
  pages={38562-38573},
  doi={10.1109/ACCESS.2025.3546017}
}

@INPROCEEDINGS{Jain_automated_2025,
  author={Jain, Eshika and Aneja, Aseem},
  booktitle={2025 International Conference on Electronics and Renewable Systems (ICEARS)}, 
  title={Automated Detection and Classification of Bean Leaf Diseases using InceptionV3: A Deep Learning Approach}, 
  year={2025},
  volume={},
  number={},
  pages={1890-1895},
  doi={10.1109/ICEARS64219.2025.10941547}
}

@article{abed_modern_2021,
  title={A modern deep learning framework in robot vision for automated bean leaves diseases detection},
  author={Abed, Sudad H and Al-Waisy, Alaa S and Mohammed, Hussam J and Al-Fahdawi, Shumoos},
  journal={International Journal of Intelligent Robotics and Applications},
  volume={5},
  number={2},
  pages={235--251},
  year={2021},
  publisher={Springer}
}

@article{singh_classification_2023,
  title={Classification of beans leaf diseases using fine tuned cnn model},
  author={Singh, Vimal and Chug, Anuradha and Singh, Amit Prakash},
  journal={Procedia Computer Science},
  volume={218},
  pages={348--356},
  year={2023},
  publisher={Elsevier}
}

@inproceedings{sunyoto_innovative_2024,
  title={Innovative Solutions for Bean Leaf Disease Detection Using Deep Learning},
  author={Sunyoto, Andi and Ariatmanto, Dhani and others},
  booktitle={2024 IEEE International Conference on Artificial Intelligence and Mechatronics Systems (AIMS)},
  pages={1--5},
  year={2024},
  organization={IEEE}
}

@article{onler_feature_2023,
  title={Feature fusion based artificial neural network model for disease detection of bean leaves.},
  author={{\"O}nler, Eray},
  journal={Electronic Research Archive},
  volume={31},
  number={5},
  year={2023}
}

@inproceedings{tan_efficientnet_2019,
  title={Efficientnet: Rethinking model scaling for convolutional neural networks},
  author={Tan, Mingxing and Le, Quoc},
  booktitle={International conference on machine learning},
  pages={6105--6114},
  year={2019},
  organization={PMLR}
}

@article{kingma2014adam,
  title={Adam: A method for stochastic optimization},
  author={Kingma, Diederik P and Ba, Jimmy},
  journal={arXiv preprint arXiv:1412.6980},
  year={2014}
}

@INPROCEEDINGS{rahunathan2023,
  author={Rahunathan, L. and Sivabalaselvamani, D. and Elakkiya, E.S. and Madhumitha, M. and Kumaresh, K.},
  booktitle={2023 3rd International Conference on Smart Data Intelligence (ICSMDI)}, 
  title={Recognition of Bean Leaf Diseases Using Neural Network and Machine Learning Techniques}, 
  year={2023},
  volume={},
  number={},
  pages={520-526},
  doi={10.1109/ICSMDI57622.2023.00098}}

@Article{sym11070939,
AUTHOR = {Arsenovic, Marko and Karanovic, Mirjana and Sladojevic, Srdjan and Anderla, Andras and Stefanovic, Darko},
TITLE = {Solving Current Limitations of Deep Learning Based Approaches for Plant Disease Detection},
JOURNAL = {Symmetry},
VOLUME = {11},
YEAR = {2019},
NUMBER = {7},
ARTICLE-NUMBER = {939},
URL = {https://www.mdpi.com/2073-8994/11/7/939},
DOI = {10.3390/sym11070939}
}

@INPROCEEDINGS{taylor2017improvingdeeplearningusing,
  author={Taylor, Luke and Nitschke, Geoff},
  booktitle={2018 IEEE Symposium Series on Computational Intelligence (SSCI)}, 
  title={Improving Deep Learning with Generic Data Augmentation}, 
  year={2018},
  volume={},
  number={},
  pages={1542-1547},
  doi={10.1109/SSCI.2018.8628742}
}

@Article{shorten_khoshgoftaar2019,
AUTHOR = {Connor Shorten and Taghi M. Khoshgoftaar},
TITLE = { A survey on Image Data Augmentation for Deep Learning},
JOURNAL = { Journal of Big Data},
VOLUME = {6},
YEAR = {2019},
NUMBER = {60},
ARTICLE-NUMBER = {},
DOI = { https://doi.org/10.1186/s40537-019-0197-0}
}

@article{10.1162/neco.1997.9.8.1735,
    author = {Hochreiter, Sepp and Schmidhuber, Jürgen},
    title = {Long Short-Term Memory},
    journal = {Neural Computation},
    volume = {9},
    number = {8},
    pages = {1735-1780},
    year = {1997},
    month = {11},
    issn = {0899-7667},
    doi = {10.1162/neco.1997.9.8.1735},
    url = {https://doi.org/10.1162/neco.1997.9.8.1735},
    eprint = {https://direct.mit.edu/neco/article-pdf/9/8/1735/813796/neco.1997.9.8.1735.pdf},
}

@article{HAQUE2025127743,
title = {An enhanced vision transformer network for efficient and accurate crop disease detection},
journal = {Expert Systems with Applications},
volume = {283},
pages = {127743},
year = {2025},
issn = {0957-4174},
doi = {https://doi.org/10.1016/j.eswa.2025.127743},
url = {https://www.sciencedirect.com/science/article/pii/S095741742501365X},
author = {Md. Ashraful Haque and Chandan Kumar Deb and Pushkar Gole and Sayantani Karmakar and Akshay Dheeraj and Mehraj Ul {Din Shah} and Subrata Dutta and M. K. Prasanna Kumar and Sudeep Marwaha}
}

\end{document}